\documentclass{article}
\usepackage{PRIMEarxiv}
\usepackage[utf8]{inputenc}
\usepackage[T1]{fontenc}
\usepackage{amsmath,amssymb,amsfonts}
\usepackage{algorithmic}
\usepackage{graphicx}
\usepackage{textcomp}
\usepackage{xcolor}
\usepackage{hyperref}
\hypersetup{hidelinks}
\usepackage{booktabs}
\usepackage{multirow}
\usepackage[ruled,vlined,linesnumbered]{algorithm2e}
\usepackage{float}
\usepackage{url}

\usepackage{tabularx}

\def\BibTeX{{\rm B\kern-.05em{\sc i\kern-.025em b}\kern-.08em
    T\kern-.1667em\lower.7ex\hbox{E}\kern-.125emX}}
\usepackage[numbers,sort&compress]{natbib}
\usepackage{microtype}
\graphicspath{{media/}}

\title{Evolving in the Agent Jungle \\
via History-Informed Opponent Awareness}
\author{
  Zhaofeng Zhang\textsuperscript{1,4,\textdagger}\quad
  Linhan Xia\textsuperscript{2,6,\textdagger}\quad
  Rui Liu\textsuperscript{3,\textdagger}\quad
  Yihao Wang\textsuperscript{5}\quad
  Binrui Shen\textsuperscript{7}\quad
  Shengxin Zhu\textsuperscript{7,8}* \\
  \textsuperscript{1}University of Edinburgh \quad
  \textsuperscript{2}University of Oklahoma \quad
  \textsuperscript{3}Imperial College London \quad
  \textsuperscript{4}University of Michigan \\
  \textsuperscript{5}University of Southern California \quad
  \textsuperscript{6}Tencent \quad
  \textsuperscript{7}Beijing Normal University \\
  \textsuperscript{8}Beijing Normal--Hong Kong Baptist University \\
  \textsuperscript{\textdagger}These authors contributed equally to this work.
  *Corresponding Author.
}

\begin{document}
\maketitle

\begin{abstract}
Learning to adapt strategies through interaction is a key step toward more general and autonomous LLM agents. Existing approaches typically achieve behavioral adaptation by revising skill libraries. However, in multi-agent environments, opponents may simultaneously update their strategies, causing the environment itself to evolve continuously. Applying skill-revision methods designed for static environments in such settings therefore amounts to updating against an obsolete reference. To address this challenge, we introduce OASE (Opponent-Aware Selective Evolution), which identifies and adopts genuinely beneficial skill revisions in dynamic multi-agent environments. Specifically, OASE conducts paired comparisons between a candidate skill and the incumbent under identical conditions anchored by historical snapshots of opponent strategies, and adopts the candidate only when its estimated payoff gain exceeds an acceptance threshold. We evaluate OASE in two decision-making scenarios: first-price auctions and private-cost Cournot competition. Experimental results show that, compared with a Reflexion-style baseline, OASE achieves a lower final equilibrium distance in both environments while accepting substantially fewer skill revisions, thereby suppressing strategy changes that lack sufficient payoff support. OASE therefore replaces blind updating with evidence‑anchored selection, allowing agents to adapt stably and efficiently even as opponents continuously evolve. The project is available \href{https://github.com/Xia12121/OASE}{here}.
\end{abstract}

\keywords{Skill Evolution \and Multi-Agent Systems \and Game Theory \and Large Language Models \and Multi-Agent Systems}

\section{Introduction}\label{intro}
Large language models (LLMs) have been employed as adaptive agents that interact with external environments. Previous work has tried to improve their behavior without updating the parameters of the underlying model. Previous works include ReAct~\citep{yao2022react}, generative agents~\citep{park2023generative}, Reflexion~\citep{shinn2023reflexion}, Self-Refine~\citep{madaan2023self}, and Voyager~\citep{wang2023voyager}. These developments suggest that an LLM agent can adapt by modifying human-readable policy artifacts rather than updating a neural policy through gradient descent. Such artifacts include prompts~\citep{guo2024connecting,fernando2023promptbreeder}, programs~\citep{ma2024eureka}, memories, and skill libraries~\citep{wang2023voyager,tziafas2024lifelong}. In single-agent settings, when the evaluation environment remains stable, a candidate revision can be assessed under approximately unchanged conditions, making adaptation relatively straightforward under stable evaluation conditions.

\begin{figure}
    \centering
    \includegraphics[width=0.7\linewidth]{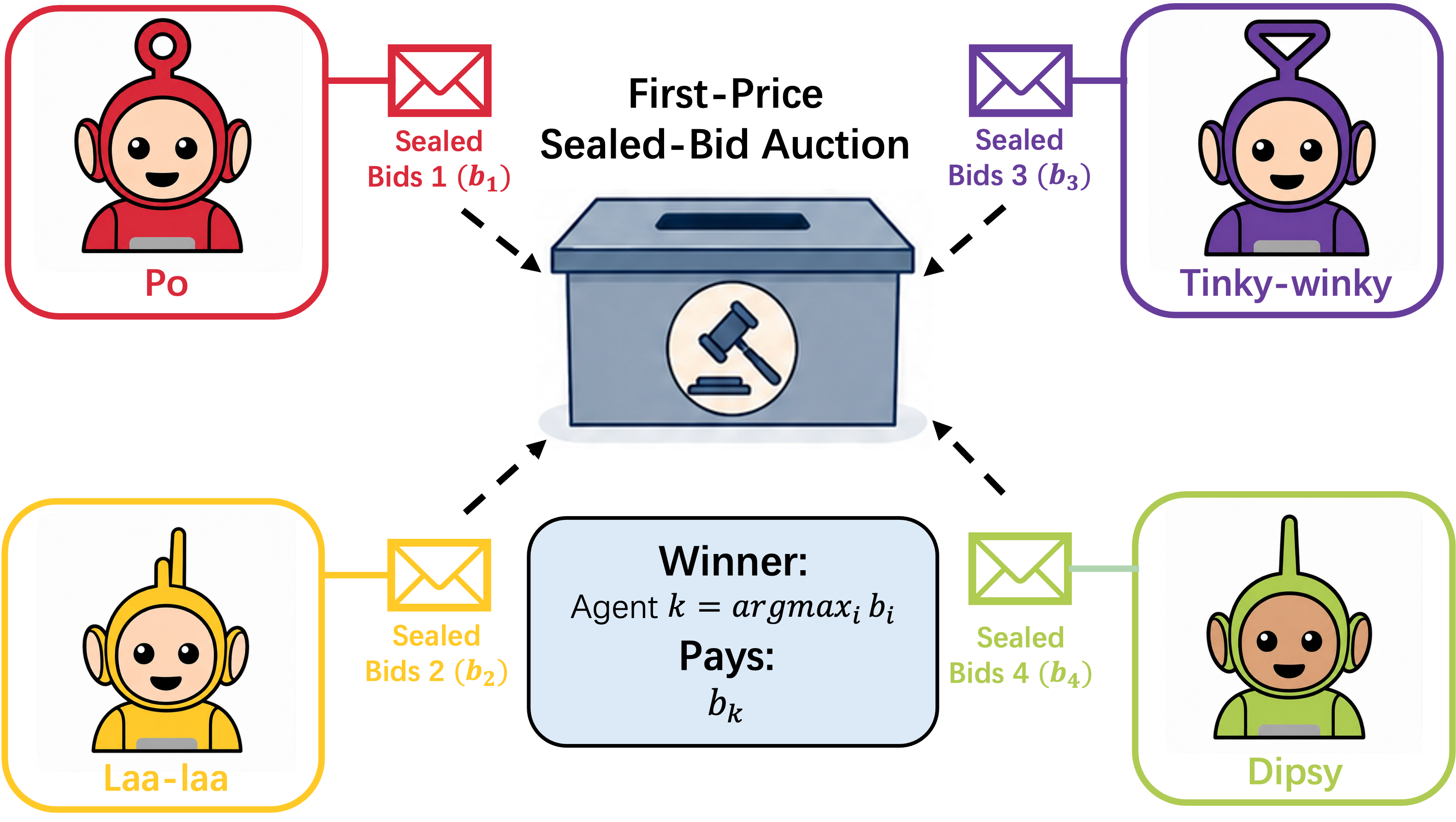}
    \caption{An example of the four-agent first-price sealed-bid auction. Each agent $i$ privately observes a valuation $v_i$ and submits a sealed bid $b_i$. The highest bidder wins, pays its own bid, and receives utility $u_i=(v_i-b_i)\mathbf{1}_{\{i\text{ wins}\}}$. Because each agent's payoff depends on the bids of the other agents, those agents form part of its effective environment.}
    \label{fig:placeholder}
\end{figure}

This adaptation becomes difficult when multiple agents revise their skill libraries simultaneously in a strategic environment, such as the first-price sealed-bid auction shown in Fig.~\ref{fig:placeholder}. From the perspective of a focal agent, the other agents constitute a part of its environment. In markets and games, an agent's payoff depends on its own strategy and the strategies adopted by the other agents. As these opponents adapt, the effective environment faced by each learner changes over time, creating a moving evaluation target in multi-agent learning~\citep{lanctot2017unified}. Strategic relationships may also be non-transitive. A strategy performing well against one opponent need not perform well against another~\citep{balduzzi2019open}. This makes it hard to apply skill-revision methods for stable environments in multi-agent settings.

This creates a revision-level credit assignment problem for LLM adaptation~\citep{minsky1961steps}. The improvement of an agent may result from various factors, such as the revision itself, opponent population, or environmental randomness. So an effective revision may appear unsuccessful when evaluated against opponents that have simultaneously become stronger. Raw trajectory feedback may reward spurious edits or reject genuinely beneficial ones. Although counterfactual methods in multi-agent reinforcement learning isolate action-level contributions under controlled conditions~\citep{foerster2018counterfactual}, they do not determine whether a complete skill revision improves upon its incumbent under matched strategic conditions.

A controlled candidate-incumbent comparison also requires a suitable opponent reference. Evaluating a candidate only against the latest population may make it appear useful because it exploits a transient opponent configuration, even though it performs poorly against strategies encountered earlier. Fictitious play and Policy Space Response Oracles (PSRO) motivate evaluating responses against distributions of previously observed strategies~\citep{lanctot2017unified}. Holding such an opponent distribution fixed during candidate evaluation provides a temporarily stable strategic target for the focal agent and reduces dependence on a single, potentially transient population state. Language-agent methods can generate interpretable strategy revisions, while population-based game-solving methods provide meaningful opponent references. Yet neither provides a controlled mechanism for deciding whether a textual revision genuinely improves upon its incumbent under co-evolving opponents.

To address this challenge, we introduce \textbf{OASE} (\textbf{O}pponent-\textbf{A}ware \textbf{S}elective \textbf{E}volution), a controlled skill-evolution framework that extends LLM-agent adaptation from stable evaluation settings to dynamically evolving multi-agent environments. OASE treats each proposed revision as a hypothesis and compares the candidate with the incumbent under matched conditions, using identical environmental randomness and opponent strategies sampled from a historical snapshot archive.  Across first-price auctions and private-cost Cournot competition, OASE achieves lower final equilibrium distance while accepting substantially fewer revisions than the baseline. Our main contributions are as follows:
\begin{itemize}
    \item We formulate LLM agents' adaptation as a revision-level credit-assignment problem. Different from previous works in stable environments, we apply skill-revision methods in \textbf{dynamic, multi-agent} environments.

    \item We introduce \textbf{OASE}, a controlled framework that determines whether a candidate skill revision really produces improvement in dynamic multi-agent environments. The candidate revision is adopted only when its estimated improvement passes an acceptance gate.

    \item  Through \textbf{paired evaluation} with \textbf{anchored snapshots} of opponent strategies, OASE estimates candidate skill revision with incumbent revision. We evaluate OASE in first-price auctions and private-cost Cournot competition. Compared with a Reflexion-style baseline, OASE achieves lower final equilibrium distance in both environments while accepting substantially fewer skill revisions.
\end{itemize}

\section{Related Work}\label{related}

\textbf{Skill Evolution in LLM Agents.} LLM-based agents can adapt through feedback and skill reuse. Related methods treat natural-language or programmatic artifacts as explicit search variables. OPRO performs black-box optimization through iterative prompting~\citep{yang2024large}, EvoPrompt combines LLM operators with evolutionary prompt search~\citep{guo2024connecting}, and PromptBreeder evolves both task prompts and mutation prompts~\citep{fernando2023promptbreeder}. Eureka similarly searches over LLM-generated reward programs~\citep{ma2024eureka}. These approaches establish that LLMs can generate and refine editable policy artifacts.

Most existing methods focus primarily on how revisions are generated and evaluated through task outcomes. Such evaluation is relatively straightforward when the relevant environment remains stable, but becomes unreliable when other adaptive agents continuously change the strategic conditions. OASE addresses this evaluation stage: rather than introducing another revision generator, it determines whether an editor-generated skill revision genuinely improves upon its incumbent under co-evolving opponents.

\textbf{Credit Assignment and Paired Evaluation.}
Credit assignment seeks to separate an individual decision maker's contribution from a joint outcome. Counterfactual multi-agent (COMA)~\citep{foerster2018counterfactual} constructs an action-level counterfactual baseline by comparing an executed action with alternative actions while holding the other agents' actions fixed through a centralized critic. OASE applies the principle to a different object and at a different stage of learning. It compares an entire candidate skill library with its incumbent through simulated rollouts, rather than comparing individual actions through a learned critic. The candidate and incumbent are evaluated under the same opponent snapshot and environmental seed, producing a paired estimate of their payoff difference. This design is also related to common-random-number evaluation in stochastic simulation. When the paired outcomes are positively correlated, shared randomness reduces the variance of their estimated difference. 

\textbf{Opponent Referencing in Dynamic Multi-agent Learning.} When multiple policies update simultaneously, each learner faces a moving strategic target. Prior work addresses such dynamics through modified optimization procedures, including symplectic gradient adjustment~\citep{balduzzi2018mechanics}, opponent shaping~\citep{letcherstable}, and competitive gradient descent~\citep{schaefer2019competitive}. These methods generally require differentiable policy parameterizations and access to gradient or higher-order information, assumptions that do not hold for discrete natural-language skill libraries.

Fictitious play responds to changing opponents through their empirical behavioral history rather than only their latest strategy~\citep{robinson1951iterative}. PSRO generalizes this principle by maintaining sets of previously discovered policies and computing response targets from a meta-strategy~\citep{lanctot2017unified}. Subsequent variants improve response computation, scalability, and strategic exploration~\citep{mcaleer2020pipeline,mcaleer2021xdo,bighashdel2024policy}.

In our work, OASE adopts the historical-opponent principle to construct a temporarily stable evaluation reference for skill revisions. An editor generates a finite local set of candidate skill libraries, and OASE compares each candidate with its incumbent against sampled historical opponent snapshots. The resulting update is therefore a controlled rather than an approximate global best response.

\textbf{LLMs in Strategic Environments.} LLMs do not necessarily present consistent game-theoretic reasoning. \citet{fan2024can} decompose game-theoretic rationality into preference formation, belief inference under uncertainty, and action selection. Their result shows that LLMs may fail to recover strategic beliefs or act consistently on beliefs they previously inferred. Fluent strategy explanations therefore do not necessarily imply rational or equilibrium-aligned behavior.

Auctions provide controlled environments in which actions depend on private information and strategic competition. Classical results characterize equilibrium bidding in private-value auctions~\citep{krishna2009auction}. Private-cost Cournot competition provides a complementary continuous-action setting with type-dependent equilibrium quantity rules~\citep{hurkens2014bayesian}. We use these two environments in the experiments because they provide external behavioral benchmarks against which induced LLM policies can be evaluated. The purpose is to measure whether textual revisions move deployed behavior toward or away from known strategic targets, rather than to establish general rationality or equilibrium convergence.

\begin{figure}[t]
  \centering
  \includegraphics[width=\textwidth]{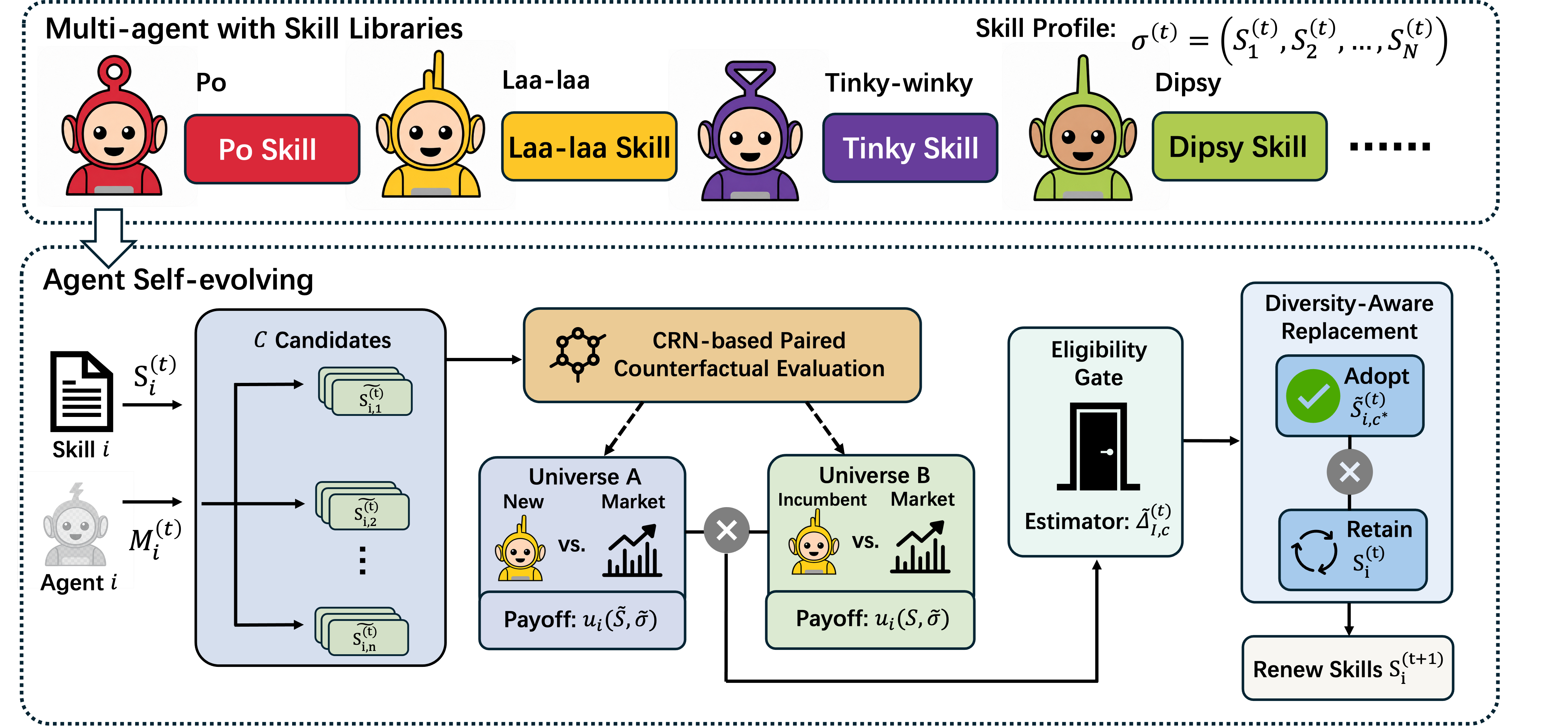}
  \caption{
  Overview of OASE. At generation $t$, an editor proposes multiple revisions to agent $i$'s incumbent skill library. Candidate and incumbent libraries are evaluated under matched seeds and anchored opponent snapshots, producing paired payoff estimates $\widehat\Delta_{i,c}^{(t)}$. Candidates that pass the payoff gate are ranked using a diversity-aware score. The highest-ranked candidate is adopted, or the incumbent is retained when no candidate qualifies. Updates are committed simultaneously and the resulting population is added to the snapshot archive.
  }
  \label{OASE_overview}
\end{figure}

\section{Methodology}\label{method}

\subsection{Overview}

Figure~\ref{OASE_overview} shows the framework of OASE. OASE extends skill adaptation to multi-agent environments by separating the generation of a skill revision from the evidence required for its adoption. Although the population evolves across generations, the snapshot archive and all evaluation conditions are frozen within each generation. This creates a temporarily stable reference against which candidate and incumbent skill libraries can be compared. The framework consists of three stages: \texttt{proposal}, \texttt{evaluation}, and \texttt{selective replacement}.

At generation $t$, an editor LLM proposes several local revisions to each focal agent's incumbent skill library. Each candidate is compared with the incumbent through paired simulations under the same environment, the same focal-agent contextual state, and the same historical opponent snapshot. Archived opponents are instantiated using their historical skill libraries together with a fixed neutral memory to prevent archived episodic memories from introducing an additional source of variation. Candidates whose one-sided confidence lower bounds do not exceed the practical improvement margin are rejected. The eligible candidates are ranked using a diversity-aware score, and the highest-ranked candidate is adopted. If no candidate qualifies, the incumbent is retained. All adoption decisions are made against the same frozen archive, and accepted updates are committed simultaneously only after all agents have been evaluated. The resulting population is then added to the snapshot archive.

\subsection{Problem Formulation}
Consider a strategic environment with $N$ learning agents. Agent $i$ maintains an individual skill library $S_i^{(t)}$ and episodic memory $M_i^{(t)}$ at generation $t$. Given private observation $o_i^{(t)}$ with public context $h^{(t)}$, its action is produced by 

\begin{equation}
    a_i^{(t)}=\pi_{\mathrm{actor}}\left(o_i;h^{(t)}, S_i^{(t)},M_i^{(t)}\right).
\end{equation}

Then the joint skill profile \(\sigma^{(t)}\) at generation $t$ is denoted by:

\begin{equation}
    \sigma^{(t)} = \left(S_1^{(t)}, S_2^{(t)}, \ldots, S_N^{(t)}\right).
\end{equation}

Let $\overline M$ denote a fixed neutral memory used to instantiate archived opponents during candidate evaluation. The corresponding historical evaluation snapshot is 

\begin{equation}
    \omega^{(t)} = \left( \left(S_1^{(t)},\overline M\right), \ldots, \left(S_N^{(t)},\overline M\right) \right).
\end{equation}

Thus, a historical snapshot preserves each opponent's historical skill library. An episode is determined by the environment mechanism $\mathcal G$ and a random seed $\xi$, which controls private information, tie-breaking, and other stochastic outcomes. When agent $i$ uses skill library $S$ against opponent profile $\sigma_{-i}$, its payoff is denoted by $u_i(S,\sigma_{-i};\xi).$

OASE performs finite local search over the textual candidates generated by the editor. Its objective at each generation is not to guarantee a globally optimal response, but to identify whether any proposed candidate exhibits sufficient positive payoff evidence relative to the incumbent under a specified opponent distribution.

\subsection{Agent Representation}

\textbf{Skill Library.} Each agent maintains a skill library 

\begin{equation}
    S_i^{(t)}=\left\{s_{i,1}^{(t)},\ldots,s_{i,K_i}^{(t)}\right\}, K_i\leq K_s
\end{equation}

where $K_s$ denotes the maximum skill-library. Each skill is a structured text object $s_{i,k}$, containing a decision condition and a corresponding strategic rule. The field \texttt{strategy} contains a decision rule. The skill library is serialized into the actor prompt. Therefore, the strategy is not executed as code; instead, the LLM actor interprets and follows the textual strategy.

The initial auction library includes legal bidding constraints and a naive shading rule, such as $b_i=0.9v_i$. This initialization differs from the analytical benchmark $b^*(v)=0.75v$, allowing us to measure whether the induced bidding behavior moves toward or away from the benchmark.

\textbf{Episodic Memory.} Each agent maintains a private episodic memory $M_i^{(t)}$, represented as a first-in, first-out queue with horizon $H$: 

\begin{equation}
    M_i^{(t)} = \left( m_i^{(\max\{0,t-H+1\})}, \ldots, m_i^{(t)} \right).
\end{equation}

After each generation, a memory-writer LLM summarizes the agent's own trajectory, including its private type, action, outcome, payoff, and relevant public market statistics. The memory is private: agent $i$ observes only its own historical outcomes and publicly available information. The skill library is the object modified by the editor, whereas episodic memory is a generation-dependent contextual state. Within each paired candidate--incumbent comparison, the focal agent uses the same memory in both rollouts. Archived opponents instead use the fixed neutral memory $\overline M$.

\textbf{Actor Policy.} The actor produces an environment-specific action according to $a_i^{(t)}$. The actor is decoded greedily with temperature zero. Invalid outputs are projected onto the corresponding feasible action set. In the first-price auction, an invalid bid is clipped to the interval $[0,v_i]$, where $v_i$ is the bidder's private valuation.

\textbf{Editor as a Local Search Operator.} At each generation, an editor LLM proposes $C$ candidate modifications to the current skill library. We write the candidate skill libraries for agent $i$ as $\widetilde{\mathcal S}_i^{(t)}$, which denotes as:

\begin{equation}
    \widetilde{\mathcal S}_i^{(t)}=\left\{\widetilde S_{i,1}^{(t)},\ldots,\widetilde S_{i,C}^{(t)}\right\}=\mathcal B_{\mathrm{edit}}\left(S_i^{(t)},M_i^{(t)}\right).
\end{equation}

Each candidate is produced through one or more bounded \textsc{Add}, \textsc{Edit}, or \textsc{Delete} operations, subject to the capacity constraint $|\widetilde S_{i,c}^{(t)}|\leq K_s$. Thus, OASE performs local search in a human-readable strategy space rather than gradient-based optimization in neural-parameter space.

\subsection{Anchored Skill Evolution} 

\textbf{Paired evaluation.} For candidate $\widetilde S_{i,c}^{(t)}$, OASE draws $m$ historical evaluation snapshots $\widehat\omega_k\sim q^{(t)}$, $k=1,\ldots,m$. For every pair, the candidate and incumbent share the same opponent snapshot $\widehat\omega_{k,-i}$, environmental random seed $\xi_k$, and focal-agent memory $M_i^{(t)}$. The paired improvement estimator is 
\begin{equation} 
\widehat\Delta_{i,c}^{(t)} = \frac{1}{m} \sum_{k=1}^{m} \left[ u_i \left( \widetilde S_{i,c}^{(t)}, M_i^{(t)}; \widehat\omega_{k,-i}; \xi_k \right) \\ - u_i \left( S_i^{(t)}, M_i^{(t)}; \widehat\omega_{k,i}; \xi_k \right) \right]. \quad
\label{eq:paired-improvement} 
\end{equation} 
Define the anchor-conditioned objective, we have:
\begin{equation} 
J_i\left(S;q^{(t)},M_i^{(t)}\right) = \mathbb E_{\widehat\omega\sim q^{(t)},\,\xi} \left[ u_i \left( S,M_i^{(t)}; \widehat\omega_{-i}; \xi \right) \right]. 
\label{payoff} 
\end{equation} 
Conditional on the generated candidate, the focal memory, and the anchor distribution, 
\begin{equation}
    \mathbb E \left[ \widehat\Delta_{i,c}^{(t)} \mid \widetilde S_{i,c}^{(t)}, M_i^{(t)}, q^{(t)} \right] \nonumber= J_i \left( \widetilde S_{i,c}^{(t)}; q^{(t)}, M_i^{(t)} \right) - J_i \left( S_i^{(t)}; q^{(t)}, M_i^{(t)} \right). 
\end{equation}

Thus, for any fixed candidate, the estimator is unbiased for its candidate--incumbent payoff difference under the specified anchor distribution. The matched opponent snapshot, focal memory, and environmental randomness do not vary between the two policies within a paired comparison, although finite-sample uncertainty remains. Let
\begin{equation}
    X_k = u_i \left( \widetilde S_{i,c}^{(t)}, M_i^{(t)}; \widehat\omega_{k,-i}; \xi_k \right), \quad
    Y_k = u_i \left( S_i^{(t)}, M_i^{(t)}; \widehat\omega_{k,-i}; \xi_k \right).    
\end{equation}

For independent pairs, the variance is:
\begin{equation}
    \operatorname{Var}
\left(
\widehat\Delta_{i,c}^{(t)}
\right)
=
\frac{1}{m}
\left[
\operatorname{Var}(X)
+
\operatorname{Var}(Y)
-
2\operatorname{Cov}(X,Y)
\right].
\end{equation}
Relative to an unpaired estimator, common random numbers reduce variance
whenever $\operatorname{Cov}(X,Y)>0$. Positive covariance is expected
when candidate and incumbent induce related behavior under the same
market realization, and we examine this condition empirically.

\textbf{Anchored Opponent Distribution.} Let $\mathcal{A}^{(t)}$ denote the archive of previous skill profiles: $\left\{ \omega^{(0)}, \omega^{(1)}, \ldots, \omega^{(t)} \right\}$. The default anchor distribution is uniform over the most recent $W$ snapshots: 
\begin{equation} 
q^{(t)} = \operatorname{Uniform} \left( \left\{ \omega^{(\max\{0,t-W+1\})}, \ldots, \omega^{(t)} \right\} \right). 
\end{equation} 
The archive and the distribution $q^{(t)}$ remain frozen throughout generation $t$. Consequently, all candidate--incumbent comparisons in that generation are conducted against the same historical reference distribution. Using a recent mixture reduces the dependence of the evaluation target on a single latest population profile. This construction is inspired by the historical-opponent principle of fictitious play and PSRO, but OASE neither constructs a restricted meta-game nor computes a meta-strategy. Because the editor proposes only a finite candidate set, the resulting procedure performs local response evaluation within an accessible textual neighborhood.

\begin{figure*}[t]
    \centering
    \includegraphics[width=\textwidth]{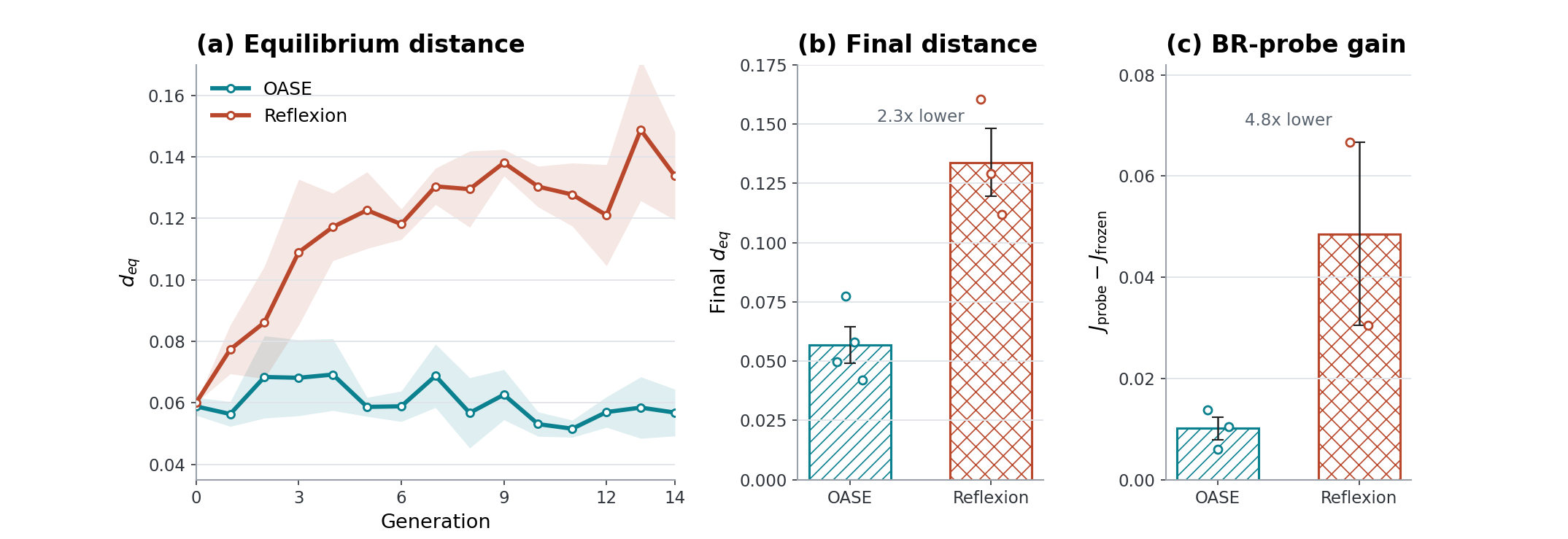}
    \caption{
    First-price auction results. 
    (a) OASE maintains a low equilibrium distance over 15 generations, while Reflexion drifts away from the symmetric BNE $b^*(v)=0.75v$.
    (b) OASE reduces final equilibrium distance by $2.3\times$ that of OASE..
    (c) Using all runs without excluding any estimates, the mean empirical BR-probe gain is $0.0101$ for OASE and $0.0486$ for Reflexion.
    Shaded regions and error bars denote standard errors across runs.
    }
    \label{fig:first-price-main}
\end{figure*}

\textbf{Gated and Diversity-Aware Replacement.} A candidate must first satisfy a payoff-improvement requirement. Under the fixed-margin gate, the eligible set $\mathcal Q_i^{(t)}$ is
\begin{equation}
  \mathcal Q_i^{(t)}=\{c\in\{1,\ldots,C\}:\widehat\Delta_{i,c}^{(t)}\geq\delta\}.  
\end{equation}

If $\mathcal Q_i^{(t)}=\varnothing$, the incumbent is retained. For each candidate, OASE computes textual novelty relative to the other
incumbent libraries:
\begin{equation}
\nu_{i,c}^{(t)}=
\frac{1}{N-1}\sum_{j\neq i}
\left[
1-
\cos
\left(
e(\widetilde S_{i,c}^{(t)}),
e(S_j^{(t)})
\right)
\right],
\end{equation}
where $e(\cdot)$ is the frozen \texttt{sentence-transformers/all-MiniLM-L6-v2} text encoder. Among eligible candidates,
OASE uses:
\begin{equation}
  G_{i,c}^{(t)}=z
\left(
\widehat\Delta_{i,c}^{(t)}
\right)
+
\lambda
z
\left(
\nu_{i,c}^{(t)}
\right)  
\label{ranking}
\end{equation}
as an auxiliary ranking score and selects $c_i^*=\arg\max G_{i,c}^{(t)}, c\in\mathcal Q_i^{(t)}.$ For each scalar candidate $\{x_{i,c}\}_{c=1}^{C}$, $z(x_{i,c})$ denotes its standardization over all $C$ candidates. If the candidate-wise standard deviation is zero, we set $z(x_{i,c})=0$. The replacement rule is
\begin{equation}
    S_i^{(t+1)}=
    \begin{cases}
\widetilde S_{i,c_i^*}^{(t)},
&
\mathcal Q_i^{(t)}\neq\varnothing,
\\
S_i^{(t)},
&
\mathcal Q_i^{(t)}=\varnothing.
\end{cases}
\end{equation}

\textbf{Significance-Adjusted Gate.} A significance-adjusted variant uses the empirical standard error of the paired differences. To account for uncertainty in the paired payoff estimate, define the per-seed difference:
\begin{equation}
D_{i,c,k}=u_i
\left(
\widetilde S_{i,c}^{(t)},
\widehat\omega_{k,-i};
\xi_k
\right)
-
u_i
\left(
S_i^{(t)},
\widehat\omega_{k,-i};
\xi_k
\right).
\end{equation}

Its empirical standard error is 

\begin{equation}
    \widehat{\mathrm{SE}_{i,c}}=
\frac{s(D_{i,c,1},\ldots,D_{i,c,m})}{\sqrt m},
\end{equation}

where $s(\cdot)$ denotes the sample standard deviation. The candidate-specific threshold is 

\begin{equation}
    \tau_{i,c}
=\max
\{
\delta+z_{1-\alpha}\widehat{\mathrm{SE}_{i,c}}
\},
\end{equation}
where $z_{1-\alpha}$ is the upper one-sided standard-normal critical
value. The corresponding eligible set is 

\begin{equation}
    \mathcal Q_{i,\mathrm{sig}}^{(t)}
=
\{
c:
\widehat\Delta_{i,c}^{(t)}
\geq
\delta\tau_{i,c}
\}.
\end{equation}

Eligible candidates are ranked using Eq.~\eqref{ranking}.
This gate is a normal-approximation heuristic for conservative
replacement rather than a finite-sample significance guarantee.

\begin{figure*}[t]
    \centering
    \includegraphics[width=\textwidth]{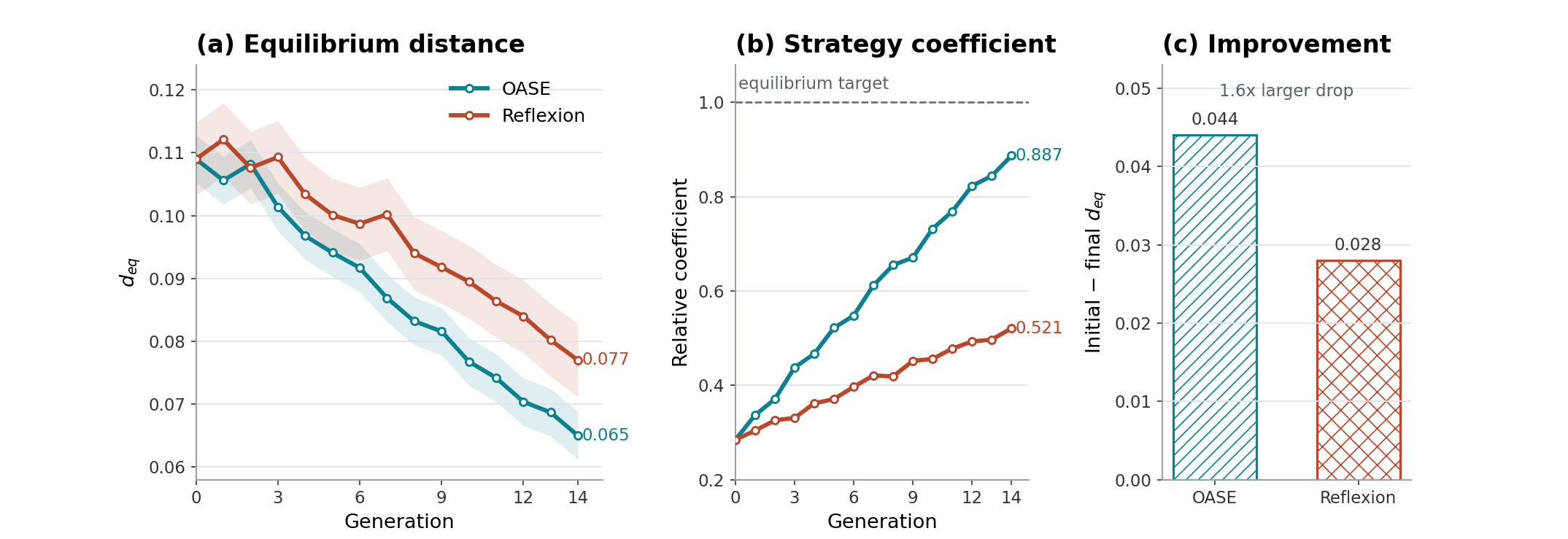}
    \caption{
    Cournot competition results based on the observed generation-level measurements. Panels (a) and (b) show the measured equilibrium-distance and strategy-coefficient trajectories, respectively. OASE ends at $d_{\mathrm{eq}}=0.065$, compared with $0.077$ for Reflexion. The strategy-coefficient trajectory ends at $0.887$ for OASE and $0.521$ for Reflexion, relative to the normalized equilibrium target of $1.0$. Panel (c) shows the measured aggregate reductions of $0.044$ and $0.028$, respectively, corresponding to a $1.6\times$ larger drop for OASE.
    }
    \label{fig:cournot-main}
\end{figure*}

\section{Experiments}\label{experiment}

\subsection{Experimental Design and Settings}
\paragraph{Environments.} We evaluate OASE in two strategic environments with externally defined equilibrium benchmarks: a symmetric first-price private-value auction and private-cost Cournot competition~\cite{krishna2009auction,hurkens2014bayesian}. The auction provides a closed-form bidding benchmark, whereas the Cournot game evaluates continuous actions conditioned on private cost.

\paragraph{Methods and budgets.} We compare OASE with a Reflexion-style baseline using the same frozen actor model, editor model, initial skill libraries, population size, and generation horizon. Each run contains $N=4$ agents and lasts for 15 generations. OASE generates $C=4$ candidate libraries per agent and generation and evaluates them using $m=64$ paired conditions sampled from an anchor window of size $W=8$. Unless otherwise stated, OASE uses the fixed-margin gate with $\delta=0.005$.

The Reflexion-style baseline updates each agent from raw trajectory feedback without paired candidate-incumbent evaluation or an acceptance gate. We report the number of editor calls and generated candidate revisions for both methods to make the proposal budget explicit.

\paragraph{Implementation details.} The actor, editor, and memory writer all use the frozen \texttt{Qwen3.5-397B-A17B} model with separate role-specific prompts. The actor generates environment actions, the editor proposes bounded modifications to the skill library, and the memory writer summarizes each agent's private trajectory after a generation. Textual novelty is computed using the frozen \texttt{sentence-transformers/all-MiniLM-L6-v2} encoder. We use a skill capacity of $K_s=5$, a memory horizon of $H=6$, and a diversity weight of $\lambda=0.3$. The actor is decoded greedily with temperature zero, and the editor uses temperature $T_{\mathrm{edit}}=0$. Each generation contains 20 main-interaction episodes.

\paragraph{Metrics and statistical protocol.}
We report means and standard errors over $R=20$ independent runs. OASE and Reflexion use matched initial libraries and matched environment seeds within each run.

Equilibrium distance is evaluated on the fixed type grid $X$. For the auction, $X$ is a valuation grid contained in $[0,1]$; for Cournot competition, $X$ is a cost grid contained in the support of $F_C$. During this probe, memory and public context are fixed to a neutral context. Accordingly, we refer to the metric as contextual equilibrium distance.

For the first-price auction, we additionally report the empirical
response-probe gain
\begin{equation}
    \widehat E_{\mathrm{probe}}
=
\widehat J_{\mathrm{probe}}
-
\widehat J_{\mathrm{inc}}.
\end{equation}

The probe is optimized using iterations and evaluated against the incumbent on the same held-out seeds. This metric is a resource-bounded deviation diagnostic rather than an estimate or certificate of the exact Nash gap.

\subsection{First-Price Auction: Equilibrium Alignment}
Figure~\ref{fig:first-price-main} reports the first-price auction results. Both methods begin from the same initial skill profile. OASE remains within a relatively narrow neighborhood of the symmetric Bayesian benchmark $b^*(v)=0.75v$, whereas the Reflexion-style baseline exhibits a larger increase in equilibrium distance over the evaluated generations. At the final generation, the mean distance is $0.057$ for OASE and $0.133$ for Reflexion.

This result is best interpreted as equilibrium alignment over the evaluated horizon rather than as proof of asymptotic convergence. In particular, OASE primarily limits subsequent departures from a relatively well-aligned policy rather than driving the distance monotonically to zero.

The empirical response probe provides a complementary resource-bounded diagnostic. Its mean gain is $0.0101$ against the final OASE populations and $0.0486$ against the Reflexion populations. Thus, under the same probe budget, the response policies found against OASE obtain smaller improvements over the incumbent. Because the probe is approximate and computationally bounded, these values do not measure exact exploitability or certify Nash equilibrium.

\subsection{Cournot Competition: Continuous-Action Equilibrium Alignment}
Cournot competition evaluates whether OASE extends from scalar auction bids to continuous quantity policies conditioned on private marginal cost. The symmetric interior Bayesian equilibrium is 

\begin{equation} 
q^*(c) = \alpha^* - \beta^*c, 
\label{cournot}
\end{equation} 

where $\alpha^*$ determines the equilibrium intercept and $\beta^*=1/(2B)$ determines the sensitivity of equilibrium output to private cost. We evaluate finite-horizon alignment with this closed-form benchmark rather than claiming asymptotic convergence.

Figure~\ref{fig:cournot-main} shows that both methods reduce equilibrium distance over the evaluated generations. OASE changes from $0.109$ to $0.065$, and Reflexion ends at $0.077$. The corresponding measured reductions are $0.044$ and $0.028$, respectively. Thus, OASE exhibits a larger endpoint improvement and a lower final distance in the available runs.

We additionally fit each induced quantity policy using $\widehat q(c)$ and report the slope ratio $r_\beta=\widehat\beta/\beta^*$. Under this diagnostic, OASE reaches $0.887$ and Reflexion reaches $0.521$, relative to the implemented target of one. This result indicates closer recovery of the reported coefficient, but should not be interpreted as verification of the complete Bayesian equilibrium unless both the intercept and slope are evaluated explicitly.

\subsection{Mechanism Diagnostics}

\textbf{Update selectivity.} OASE accepts an average of $0.78$ and $0.73$ population-level edits per generation in the first-price and Cournot environments, respectively. The Reflexion-style baseline adopts $3.98$ and $4.00$ edits. Because the population contains four agents and each agent can adopt at most one revision per generation, the baseline updates nearly every agent at every generation, whereas OASE frequently retains the incumbent.

This diagnostic confirms that OASE implements selective rather than unconditional replacement. It does not by itself establish that fewer updates cause lower equilibrium distance.

\textbf{Variance of paired evaluation.} In the first-price diagnostic, the variance of the paired candidate-incumbent estimator is $5.60\times10^{-4}$, compared with $1.11\times10^{-3}$ for the corresponding unpaired estimator. The paired variance is therefore $50.3\%$ of the unpaired variance. This supports the positive-covariance condition underlying common-random-number evaluation in this environment.

\begin{table}[t]
\centering
\label{tab:mechanism}
\begin{tabular}{lcc}
\toprule
\multicolumn{3}{c}{\textbf{Update selectivity}}\\
Environment & OASE & Reflexion\\
\midrule
First-price edits / generation & 0.78 & 3.98\\
Cournot edits / generation & 0.73 & 4.00\\
\midrule
\multicolumn{3}{c}{\textbf{Estimator variance}}\\
Diagnostic & Paired CRN & Unpaired\\
\midrule
Variance of $\widehat\Delta$ &
$5.60\times10^{-4}$ & $1.11\times10^{-3}$\\
Relative variance & 50.3\% & 100\%\\
\bottomrule
\end{tabular}
\caption{
Mechanism diagnostics. Accepted edits are population-level totals per
generation for $N=4$ agents. The variance comparison is computed for
the first-price candidate--incumbent estimator.
}
\end{table}

\section{Discussion}\label{discussion}

The experiments indicate that OASE differs from trajectory-driven revision primarily in how proposed edits are evaluated and retained. The actor and editor can still produce noisy textual revisions, but OASE treats each revision as a hypothesis rather than as evidence of improvement. The empirical results are consistent with two properties of this design: fewer population updates and a lower-variance paired comparison. These diagnostics support a selective-update interpretation, although the present experiments do not isolate every OASE component.

\subsection{Selective Updating and Lower Update Churn}

OASE accepts substantially fewer revisions than the Reflexion-style baseline. This behavior follows directly from the eligibility gate: when no candidate exceeds the payoff margin, the incumbent is retained. Such retention is potentially valuable near a useful strategic rule, because editor-generated changes need not be beneficial.

However, low update frequency is not itself evidence of successful learning; an overly conservative gate could simply freeze the population. In the present experiments, lower churn occurs together with lower final equilibrium distance, a pattern consistent with the hypothesis that filtering unsupported revisions limits harmful strategy drift. A direct causal attribution would require a controlled comparison against OASE without the acceptance gate.

\subsection{Paired Evaluation and Estimation Noise}

By evaluating candidate and incumbent libraries under the same random seed and anchored opponent snapshot, OASE estimates their payoff difference using common random numbers. In the first-price diagnostic, the paired estimator has approximately half the variance of the corresponding unpaired estimator.

This result supports the proposed estimator-level mechanism: shared conditions provide a cleaner estimate of the local candidate-incumbent difference when their outcomes are positively correlated. It does not, by itself, establish that variance reduction causes the final equilibrium-distance improvement. Such a claim would require training an otherwise identical OASE variant using unpaired evaluation.

\section{Conclusion}

We introduced OASE, a framework for evaluating and selectively adopting skill revisions in co-evolving LLM populations. OASE compares candidate and incumbent libraries under matched environmental randomness and anchored opponent snapshots, and retains the incumbent when no candidate provides sufficient payoff evidence.

Across first-price auctions and private-cost Cournot competition, OASE obtains lower final equilibrium distance than a Reflexion-style baseline while accepting substantially fewer revisions. Paired common-random-number evaluation also yields lower estimator variance in the first-price diagnostic. These findings suggest that controlled candidate--incumbent evaluation can make textual strategy adaptation more selective and statistically reliable.

\section*{Acknowledgment}
This work was supported in part by the National Key Technologies Research and Development Program (2025YFG0202100; 2025YFG0202600). The authors have no competing interests to declare that are relevant to the content of this article.

\bibliographystyle{unsrt}
\bibliography{ref}

\appendix

\section{Strategy Environments}
\paragraph{First-price auction.}
We consider a symmetric first-price sealed-bid auction with $N=4$
risk-neutral bidders. Each bidder $i$ independently observes a private
valuation
\begin{equation}
v_i
\overset{\mathrm{i.i.d.}}{\sim}
\mathrm{Uniform}[0,1].
\end{equation}
Bidder $i$ submits a bid $b_i$ from the feasible set $[0,v_i]$. The bidder with
the highest bid wins and pays its submitted bid; ties are broken uniformly at
random among the highest bidders. The realized payoff is
\begin{equation}
u_i
=
\left(v_i-b_i\right)
\mathbf{1}_{
\left\{
i_\text{ wins}
\right\}}.
\end{equation}
Under independent private values, risk neutrality, and four symmetric bidders,
the symmetric Bayesian Nash equilibrium bidding function is
\begin{equation}
b^*(v)
=
\frac{N-1}{N}v
=
\frac{3}{4}v.
\label{eq:auction-bne}
\end{equation}

\paragraph{Private-cost Cournot competition.}
We consider $N=4$ risk-neutral firms that simultaneously choose nonnegative
quantities $q_i$. Total output is $Q=\sum_{j=1}^{N}q_j$, and the inverse demand
function is
\begin{equation}
P(Q)=A-BQ,
\qquad
A>0,\quad B>0.
\end{equation}
Firm $i$ privately observes its constant marginal cost $c_i$, where the costs
are independently drawn from a common distribution $F_C$ with mean
\begin{equation}
\mu_c=\mathbb E[c_i].
\end{equation}
Its profit is
\begin{equation}
u_i
=
\left(
A-BQ-c_i
\right)q_i.
\end{equation}

For an interior symmetric Bayesian Nash equilibrium, the type-dependent
quantity rule is shown in Eq.~\ref{cournot}, where

\begin{equation}
\beta^*
=
\frac{1}{2B},
\qquad
\alpha^*
=
\frac{
A+\frac{N-1}{2}\mu_c
}{
B(N+1)
}.
\label{eq:cournot-coefficients}
\end{equation}

For $N=4$, these coefficients reduce to
\begin{equation}
\beta^*
=
\frac{1}{2B},
\qquad
\alpha^*
=
\frac{
A+\frac{3}{2}\mu_c
}{
5B
}.
\end{equation}
If the affine expression becomes negative for some cost types, the
nonnegative-quantity benchmark is
\begin{equation}
q^*(c)
=
\left[
\alpha^*-\beta^*c
\right]_+
=
\max
\left\{
0,\alpha^*-\beta^*c
\right\}.
\end{equation}

\end{document}